\documentclass{sig-alternate}
\usepackage{multirow}
\usepackage{url}
\begin{document}
%
\conferenceinfo{}{}

\title{Multi-Scale Convolutional Neural Networks for Time Series Classification}

%
%
%
%
%

\numberofauthors{3} 
%
\author{
%
%
\alignauthor
Zhicheng Cui\\
       \affaddr{Department of Computer Science and Engineering}\\
       \affaddr{Washington University in St. Louis, USA}\\
       \email{z.cui@wustl.edu}
\alignauthor
Wenlin Chen\\
       \affaddr{Department of Computer Science and Engineering}\\
       \affaddr{Washington University in St. Louis, USA}\\
       \email{wenlinchen@wustl.edu}
\alignauthor
Yixin Chen\\
       \affaddr{Department of Computer Science and Engineering}\\
       \affaddr{Washington University in St. Louis, USA}\\
       \email{chen@cse.wustl.edu}
}

\newcommand{\bx}[0]{{\textbf x}}
\newcommand{\bv}[0]{{\textbf v}}
\newcommand{\mR}[0]{\mathcal{R}}
\newcommand{\mP}[0]{\mathcal{P}}
\newcommand{\name}[0]{Multi-scale Convolutional Neural Network}
\newcommand{\abbrev}[0]{MCNN}

\newdef{definition}{Definition} 
\newtheorem{theorem}{Theorem} 
\newtheorem{lemma}{Lemma} 

\maketitle
\begin{abstract}

Time series classification (TSC), the problem of predicting class labels of time series, has been around for decades within the community of data mining and machine learning, and found many important applications such as biomedical engineering and clinical prediction. However, it still remains challenging and falls short of classification accuracy and efficiency. Traditional approaches typically involve extracting discriminative features from the original time series using dynamic time warping (DTW) or shapelet transformation, based on which an off-the-shelf classifier can be applied. These methods are ad-hoc and separate the feature extraction part with the classification part, which limits their accuracy performance. Plus, most existing methods fail to take into account the fact that time series often have features at different time scales. To address these problems, we propose a novel end-to-end neural network model, \name{} (\abbrev{}), which incorporates feature extraction and classification in a single framework. Leveraging a novel multi-branch layer and learnable convolutional layers,  \abbrev{} automatically extracts features at different scales and frequencies, leading to superior feature representation. \abbrev{} is also computationally efficient, as it naturally leverages GPU computing. We conduct comprehensive empirical evaluation with various existing methods on a large number of benchmark datasets, and show that \abbrev{} advances the state-of-the-art by achieving superior accuracy performance than other leading methods.

\end{abstract}

{Data Mining}




\section{Introduction}
Our daily lives constantly produce time series data, such as stock prices, weather readings, biological observations, health monitoring data, etc. In the era of big data, there are increasing needs to extract knowledge from time series data, among which a main task is time series classification (TSC),  the problem of predicting class labels for time series. It has been a long standing problem with a large scope of real-world applications. For example, there has been active research on clinical prediction, the task of predicting whether a patient might be in danger of certain deterioration based on the patient's clinical time series such as ECG signals. A real-time deterioration warning system powered by TSC has achieved unprecedented performance compared with traditional clinical approaches and been applied in major hospitals~\cite{mao2012integrated}.

Most existing TSC approaches fall into  two categories~\cite{xing2010brief}: distance-based methods and feature-based methods.

For distance-based methods, the key part is to measure the similarity between any given two time series. Based on the similarity metrics, the classification can be done using algorithms such as k-nearest neighbors (kNN) or support vector machines (SVM) with  similarity-based kernels. The most notable similarity measurement is dynamic time warping (DTW) which aligns two time series with dynamic warping to get the best fit. It could be easily done through dynamic programming.

For feature-based methods, each time series is characterized with a feature vector and any feature-based classifier (e.g. SVM or logistic regression) can be applied to generate the classification results. There have been many hand-crafted feature extraction schemes across different applications. For example, in a clinical prediction application, each time series is divided into several consecutive windows and features are extracted from each window. The final feature vector is a concatenation of feature vectors from all windows~\cite{mao2012integrated}.
The features include simple statistics such as mean and variance, as well as complex features from detrended fluctuation analysis and spectral analysis.
Another approach extracts features based on shapelets which can be regarded as a signature subsequence of the time series. Typically, potential candidate shapelets are generated in advance and they can be used in different ways. For example, they can be considered as a dictionary and each shapelet is regarded as a word. The time series is then described by a bag-of-word model. A more recent study~\cite{lines2012shapelet} constructs the feature vector such that the value of each feature is the minimum distance between anywhere in the time series and the corresponding shapelet. A drawback of the shapelet method is that it requires extensive search for the discriminative shapelets from a large space.
To bypass the need of trying out lots of shapelet candidates, Grabocka~\emph{et al.}~\cite{grabocka2014learning} propose to jointly learn a number of shapelets of the same size along with the classifier. However, their method only offers linear separation ability.

In recent years, convolutional neural networks (CNNs) have led to impressive results in object
recognition~\cite{Krizhe12}, face verification~\cite{Facenet15}, and audio classification~\cite{Lee09}.

A key reason for the success of CNNs is its ability to automatically learn complex feature representations using its convolutional layers.
With the great recent success of deep learning and the presence of so many various handcrafted features in TSC, it is natural to ask a question: is it possible to automatically learn the feature representation from time series? However, there have not been many research efforts in the area of time series to embrace deep learning approaches. In this paper, we advocate a novel neural network architecture, \name{} (\abbrev{}), a  convolutional neural network specifically designed for classifying time series.

A distinctive feature of \abbrev{} is that its first layer contains multiple branches that perform various transformations of the time series, including those in the frequency and time domains, extracting features of different types and time scales. Subsequently, convolutional layers apply dot products between the transformed waves and 1-D learnable filters, which is a general way to automatically recognize various types of features from the input. As a single convolutional layer can detect local patterns similar to shapelets, stacking multiple convolutional layers can construct more complex patterns. As a result, \abbrev{} is a powerful general-purpose framework for TSC. Different than traditional TSC methods,  \abbrev{} is an end-to-end model without requiring any handcrafted features. We conduct comprehensive experiments and compare with many existing TSC models. Strong empirical results show that \abbrev{} elevates the  state-of-the-art of TSC. It gives superior overall performance, surpassing most existing models by a large margin, especially when enough training data is present.


\section{Multi-Scale Convolutional Neural Network (MCNN) for TSC}

In this section, we formally define the aforementioned time series classification (TSC) problem. Then we describe our \abbrev{}  framework for solving TSC problems.

\subsection{Notations and Problem Definition}

A time series is a sequence of real-valued data points with timestamps.  In this paper, we focus on time series with identical interval length. We denote a time series as $T = \{t_1, t_2, ..., t_n\}$, where $t_i$ is the value at time stamp $i$ and there are $n$ timestamps for each time series.

We denote a labelled time series dataset as $D = \{(T_i, y_i)\}_{i=1}^N$ which contains $N$ time series and their associated labels. For each $i=1, \cdots, N$, $T_i$ represents the $i^{th}$ time series  and its label is $y_i$. For ease of presentation, in this paper we consider classification problems where $y_i$ is a categorical value in $\mathcal{C}=\{1,\cdots,C\}$ where $C \in \mathbb{Z}^+$ is the number of labels. However, our framework can be easily extended to real-valued regression tasks. The TSC problem is to build a predictive model to predict a class label $y \in \mathcal{C}$ given an input time series $T$. Unlike some previous works, we do not require all training and testing time series to have the same number of timestamps in our framework.

\subsection{MCNN framework}
Time series classification is a long standing problem that has been studied for decades. However, it remains a very challenging problem despite great advancement in data mining and machine learning.  There are some key factors contributing to its difficulty. First, different time series may require feature representations at different time scales. For example, it is found that certain long-range (over a few hours involving hundreds of time stamps) patterns in body temperature time series have predictive values in forecasting sepsis~\cite{Drewry13}.
Existing TSC features  can rarely adapt to the right scales.
Second, in real-world time series data, discriminative patterns in the time series is often distorted by high-frequency perturbations and random noises. Automatic smoothing and de-noising procedures are needed to make the overall trend of the time series more clear.

\begin{figure*}[t]
    \centering
    \includegraphics[width=0.9\textwidth]{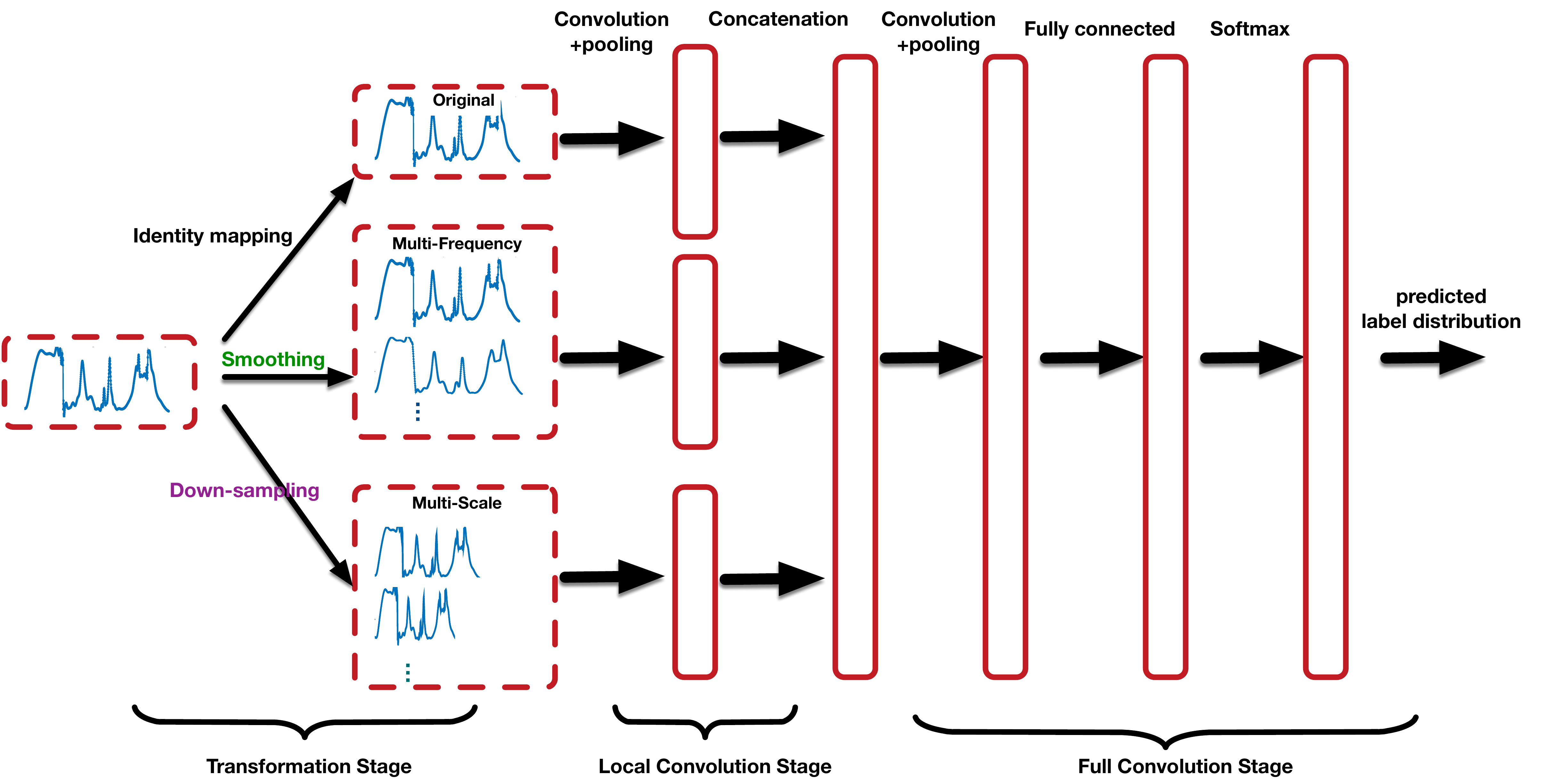}
    \caption{Overall architecture of MCNN.}
    \label{fig.mscnn-ts}
\end{figure*}

To address these problems for TSC, we propose a multi-scale convolutional neural network  (\abbrev{}) framework in which the input is the time series to be predicted and the output is its label. The overall architecture of \abbrev{} is depicted in Figure~\ref{fig.mscnn-ts}.

The \abbrev{} framework has three sequential stages: transformation, local convolution, and full convolution.

1) The \textbf{transformation stage} applies various transformations on the input time series. We currently include identity mapping, down-sampling transformations in the time domain, and spectral transformations in the frequency domain. Each part is called a \emph{branch}, as it is a branch input to the convolutional neural network.

2) In the \textbf{local convolution stage}, we use several convolutional layers to extract the features for each branch. In this stage, the convolutions for different branches are independent from each other. All the outputs will pass through a max pooling procedure with multiple sizes.

3) In the \textbf{full convolution stage}, we concatenate all extracted features and apply several more convolutional layers (each followed by max pooling), fully connected layers, and a softmax layer to generate the final output. This is an entirely end-to-end system and all parameters are trained jointly through back propagation.

\subsection{Transformation stage}

\textbf{Multi-scale branch.}
A robust TSC model should be able to capture temporal patterns at different time scales. Long-term features reflect overall trends and short-term features indicate subtle changes in local regions, both of which can be potentially crucial to the prediction quality for certain tasks.

In the multi-scale branch of \abbrev{}, we use down-sampling to generate sketches of a time series at different time scales. Suppose we have a time series $T = \{t_1, t_2, ..., t_n\}$ and the down-sampling rate is $k$, then we will only keep every $k^{th}$ data points in the new time series:
\begin{equation}
T^k = \{t_{1 + k * i}\}, \ i = 0,1,..., \lfloor \frac{n-1}{k} \rfloor.
 \end{equation}
 Using this method, we generate multiple new input time series with different down sampling rates, e.g. $k=2, 3, \cdots$.

\textbf{Multi-frequency branch.}
In  real-world applications, high-frequency perturbations and random noises widely exist in the time series data due to many reasons, which poses another challenge to achieving high prediction accuracy. It is often hard to extract useful information on raw time series data with the presence of these noises. In \abbrev{}, we adopt low frequency filters with multiple degrees of smoothness to address this problem.

A low frequency filter can reduce the variance of time series. In particular, we employ moving average to achieve this goal. Given an input time series, we generate multiple new time series with varying degrees of smoothness using moving average with different window sizes. This way, newly generated time series represent general low frequency information, which make the trend of time series more clear. Suppose the original time series is $T = \{t_1, t_2, ..., t_n\}$, the moving average works by converting this original time series into a new time series
\begin{equation}
T^\ell = {\frac{x_i + x_{i + 1} + ... + x_{i + \ell - 1}}{\ell}},
\end{equation}
where $\ell$ is the window size and $i = {0,1,...n-\ell+1}$. With different $\ell$, \abbrev{} generates multiple time series of different frequencies, all of which will be fed into the local convolutional layer for this branch. Different from the multi-scale branch, each time series in the multi-frequency branch has the same length, which allows us to assemble them into multiple channels for the following convolutional layer.

\subsection{Local convolution stage}

\textbf{Local convolution.}
After down sampling, we obtain multiple time series with different lengths from a single input time series. We apply independent 1-D local convolutions on each of these newly generated time series. In particular, the filter size of local convolution will be the same across all these time series. Note that, with a same filter size, shorter time series would get larger local receptive field in the original time series. This way, each output from the local convolution stage captures a different scale of the original time series. An advantage of this method is that, by down sampling the time series instead of increasing the filter size, we can greatly reduce the number of parameters in the local convolutional layer.

\textbf{Max pooling with multiple sizes.}
Max pooling, a form of non-linear down-sampling, is also performed between successive convolutional layers in \abbrev{}. This can reduce feature maps' size as well as the amount of following layers' parameters to avoid overfitting and improve computation efficiency. More importantly, the max pooling operation introduces invariance to spatial shifting, making \abbrev{} more robust.

Instead of using small pooling sizes like $2$ or $5$, in \abbrev{} we introduce a variable called the \emph{pooling factor}, $p$, which is the length after max pooling. Suppose the output time series after convolution has a length of $n$, then both our pooling size and stride in max pooling are $\frac{n}{p}$. The pooling size is fairly large since $p$ is often chosen from $\{2,3,5\}$. By doing this, we can have more filters and enforce each filter to learn only a local feature, since in the backpropogation phase, filters will be updated based on those few activated convolution parts.

\subsection{Full convolution stage}
After extracting feature maps from multiple branches, we concatenate all these features and feed them into other convolutional layers as well as a fully connected layer followed by a softmax transformation. Following \cite{szegedy2015going} , we adopt the technique of \emph{deep concatenation} to concatenate all the feature maps vertically.

The output of \abbrev{} will be the predicted distribution of each possible label for the input time series. To train the neural network, \abbrev{} uses the cross-entropy loss defined as:
 \begin{equation}
 \max_{\mathbb{W}, \textbf{b}} \sum_{i=1}^{N} \log o_{y_i}^{(i)},
 \end{equation}
where $o_{y_i}^{(i)}$ is the $y_i^{th}$ output of instance $i$ through the neural network, which is the probability of its true label. The parameters $\mathbb{W}$ and bias $\textbf{b}$ in \abbrev{} are those in local and full convolutional layers, as well as those in the fully connected layers, all of which are learned jointly through back propagation.

\subsection{Data augmentation}
One advantage for our framework is the ability to deal with large scale datasets. When dealing with smaller datasets, convolutional nets tend to overfit. Currently, most publicly available TSC datasets have limited sizes. To overcome this problem, we propose a data augmentation technique on the original datasets in order to avoid overfitting and improve the generalization ability. For massive datasets with abundant training data, data augmentation may not be needed.

We propose \emph{window slicing} for the data augmentation. For a time series $T = \{t_1, \cdots, t_n\}$, a slice is a snippet of the original time series, defined as $S_{i:j} = \{t_i, t_{i + 1},...,t_j\}$, $1 \le i\le j\le n$. Suppose a given time series $T$ is of length $n$, and the length of the slice is $s$, our slicing operation will generate a set of $n$-$s$+1 sliced time series:
\begin{equation}
Slicing(T,s) = \{S_{1:s}, S_{2:s+1}, \cdots, S_{n-s+1:n}\},
\end{equation}
where all the time series in $Slicing(T,s)$ have the same label as their original time series $T$ does.

We apply window slicing on all time series in a given training dataset. When doing training, all the training slices are considered independent training instances. We also do window slicing when predicting the label of a testing time series. We first use the trained MCNN to predict the label of each of its slices, and then use a majority vote among all these slices  to make the final prediction. Another advantage of slicing is that the time series are not required to have equal length since we can always cut all the time series into the same length using window slicing.


\section{Discussion}

In this section, we discuss several properties of the \abbrev{} framework and its relations to some other important works.

\subsection{Effectiveness of convolution filters}
Convolution has been a well-established method for handling sequential signals~\cite{mallat1999wavelet}. We advocate that it is also a good fit for capturing characteristics in time series. Suppose $f$ is a filter of length $m$ and $T$ is a time series. Let $T \cdot f$ be the result of 1-dimensional discrete convolution. The $i^{th}$ element of the result is given by
$$(T \cdot f)[i] = \sum_{j = 1}^{m} f_{m + 1 - j} \cdot t_{i + j - 1}$$
Depending on the filter, the convolution is capable of extracting many insightful information from the original time series. For example, if $f = [1,-1]$, the result of the convolution would be the gradient between any two neighboring points. However, is \abbrev{} able to learn such kind of filters? The answer is yes. To show this, we train \abbrev{} on a real-world dataset Gun\_Point with a filter whose size is 15 ($m=15$). For  illustration, we pick one of the filters learned by \abbrev{} as well as 3 time series from the dataset.

\begin{figure}[t]
    \centering
    \includegraphics[width=0.51\textwidth]{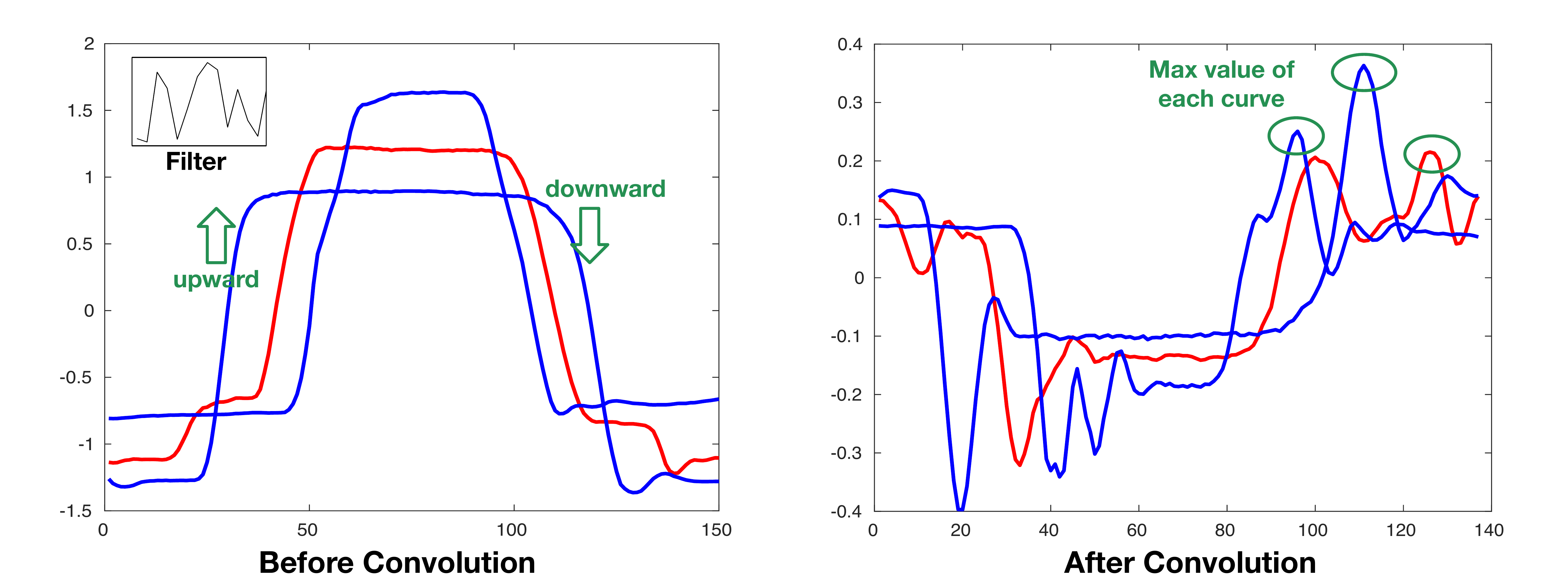}
    \caption{Three time Series on Gun\_Point dataset before and after performing the convolution operation. The two blue curves belong to one class and the red curve belongs to a different class.}
    \label{fig.conv-op}
\end{figure}

We show the shape of this selected filter and these 3 time series on the left of Figure~\ref{fig.conv-op}, and the shape after convolution with the filter. Here, the two blue curves belong to one class and the red curve belongs to a different class. The learned filter (shown in the left figure) may look random at the first glance. However, a closer examination shows that it makes sense.

First, we can observe from the left figure that each time series has a upward part and a downward part no matter which label it has. After convolution with the filter, all three new signals form a valley at the location of the upward part and a peak at the location of the downward part. Second, since in \abbrev{} we use max pooling right after convolution, the learned filter correctly finds that the downward part is more important, as max pooling only picks the maximum value from each convolved signal. As a result, the convolution and max pooling correctly differentiate the blue curves and red curve, since the maximum values of the two blue curves after convolution is greater than that of the red curve.  By this visualization, \abbrev{} also offers certain degree of interpretability as it tells us the characteristic found by \abbrev{}. Third, these three time series have similar overall shapes but different time scales, as the ``plateau" on the top have different lengths. It is very challenging for other methods such as DTW or shapelet to classify them. However, a single filter learned by \abbrev{} coupled with max pooling can classify them well.
 
\begin{figure}[t]
    \centering
    \includegraphics[width=0.3\textwidth]{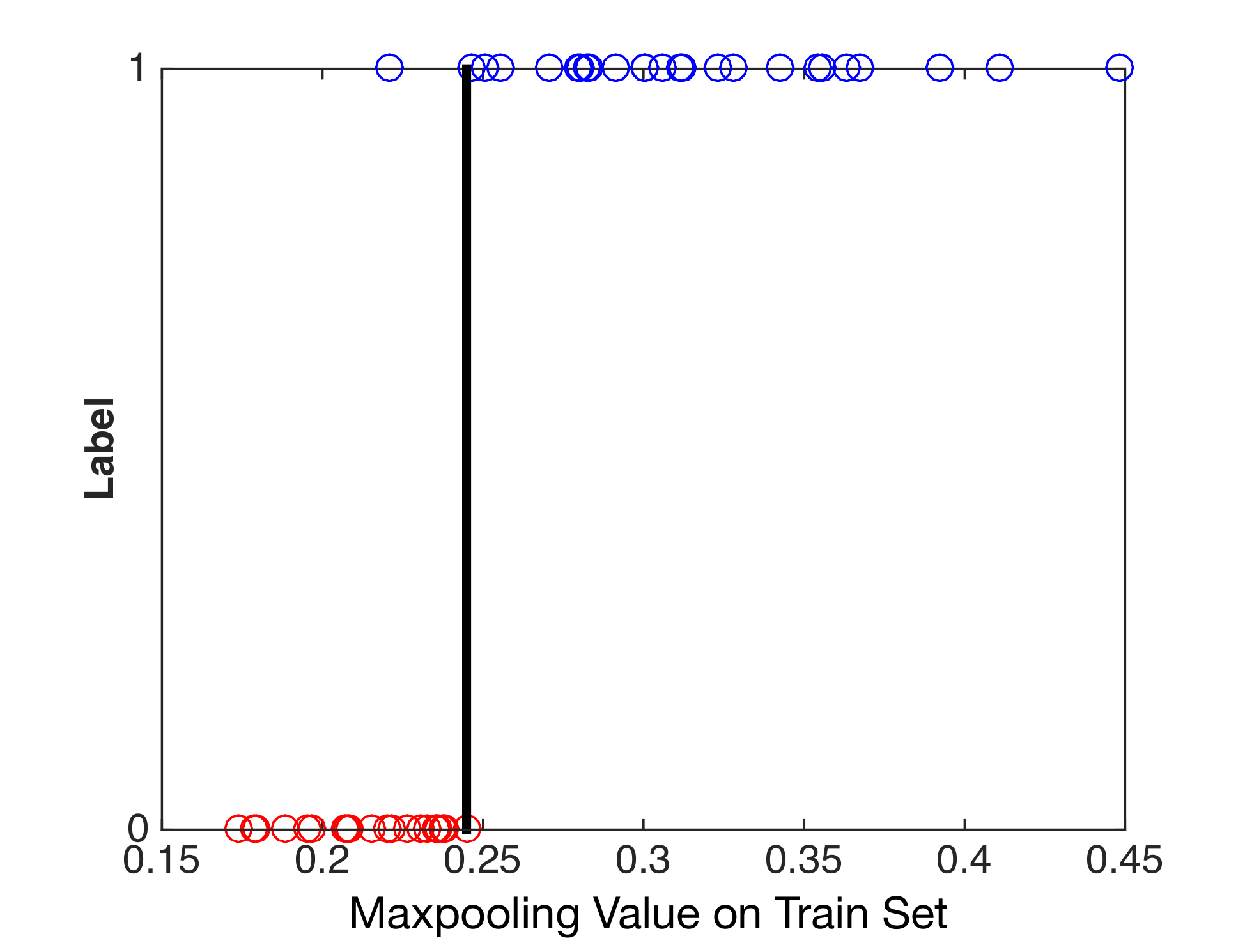}
    \caption{After training \abbrev{}, we picked one filter and perform the convolution operation and max pooling on all training data.}
    \label{fig.conv-result}
\end{figure}

To further demonstrate the power of convolution filters for TSC, we compute the max pooling result of all times series in the train set convolving with the  filter shown in Figure~\ref{fig.conv-op}, and show all of them in Figure~\ref{fig.conv-result}. Here, each point corresponds to a time series in the dataset. The blue and red points correspond to two different classes, respectively. The x-axis is the max-pooling value of each point, and y-axis is the class label. We can see from Figure~\ref{fig.conv-result} that,  if we set the classification threshold at around $0.25$, one single convolution filter can already achieve very high accuracy to classify the dataset.

\subsection{Relation to learning shapelets}
A major class of TSC methods are based on shapelet analysis which assumes that time series are featured by some common subsequences. Shapelet can either be extracted from existing time series, or learned from the data. A recent study~\cite{grabocka2014learning} proposes a learning time series shapelet (LTS) method which achieves unprecedented performance improvement over simple extraction. In the LTS method, each time series can be represented by a feature vector in which each feature is the similarity between the time series and a shapelet. A logistic regression is applied on this new representation of time series to get the final prediction. Both the shapelets and parameters in the logistic regression model are jointly learned.

There is a strong relevance between the LTS method and \abbrev{}, as both learn the parameters of the shapelets or filters jointly with a classifier. In fact, LTS can be viewed as a special case of \abbrev{}. To make this more clear, let us first consider a simpler architecture, a special case of \abbrev{}, where there is only one identity branch, and the input time series is processed by a 1-D convolutional layer followed by a softmax layer. The 1-D convolutional filter in the model can be regarded as a shapelet. The second layer (after convolution) is the new representation of the input time series. In this case, each neuron in the second layer is a inner product between the filter (or shapelet) and the corresponding window of the input time series. From this, we can see that \abbrev{} model adopts inner product as the similarity measurement while LTS employs the Euclidean distance.

To further show the relationship between inner product in convolution and Euclidean distance, we can actually express the Euclidean distance in the form of convolution. Let $T\ominus f$ be the Euclidean distances between a time series $T = \{t_1, \cdots, t_n\}$ and a filter $f = \{f_1, \cdots, f_m\}$, its $i^{th}$ element is:
 
\begin{eqnarray}
		(T\ominus f)[i] &=& \sum_{j = 1}^{m} \Biggl(t_{i+ j - 1} - f_{m + 1 - j}\Biggl)^2 \nonumber\\
		 &=&\sum_{j = 1}^{m} t^2_{i + j -1} + \sum_{j = 1}^{m} f^2_{m + 1 -j} \nonumber\\
		&&  \hphantom{{}\sum no}-2\sum\limits_{j = 1}^{m}t_{i + j - 1}f_{m + 1 -j} \nonumber \\
		 &=& \sum_{j = 1}^{m} t^2_{i + j -1} + \sum_{j = 1}^{m}f^2_j - 2 (T \cdot f)[i]
	\label{eq.euc_conv}
\end{eqnarray}
From Eq.~\eqref{eq.euc_conv}, the Euclidean distance is nothing but the combination of convolution $T \cdot f$ (after flipping the sign of $f$) and the $\ell_2$ norms of $f$ and a part of $T$. The first term in Eq.~\eqref{eq.euc_conv} is a constant for each time series, and therefore can be regarded as a bias which \abbrev{} has incorporated in the model. We can thus see that learning shapelets is a special case of learning convolution filters when the filters are restricted to have the same $\ell_2$ norm.
Moreover, if we consider the full \abbrev{} framework, its multi-scale and multi-frequency branches make it even more general to handle different time scales and noises.
 
Eq.~\eqref{eq.euc_conv} also gives us a hint on how to use convolution neural networks to implement Euclidean distances. By doing this, the Euclidean distance of between the time series and the shapelets can be efficiently computed leveraging deep learning packages and the speedups from their GPU implementation.


\section{Related Work}
TSC  has  been studied for long time. A plethora of time series classification algorithms have been proposed.  Most traditional TSC methods fall into two categories: distance based methods that use kNN classifiers on top of distance measures between time series, and feature based classifiers that extract or search for deterministic features in the time or frequency domain and then apply traditional classification algorithms. In recent years, some ensemble methods that collect many TSC classifiers together have also been studied. A full review of these methods is out of the scope here but we will do a comprehensive empirical comparison with leading TSC methods in the next section. Below, we review some works that are most related to \abbrev{}.

In recent years, there have been active research on deep neural networks~\cite{hinton2006fast,bengio2009learning,arel2010deep} that can combine hierarchical feature extraction and classification together. Extensive comparison has shown that convolution operations in CNN have better capability on extracting meaningful features than ad-hoc feature selection~\cite{martinez2013learning}. However,  applications of CNN to TSC have not been studied until recently.
 
A multi-channel CNN has been proposed to deal with multivariate time series~\cite{zheng2014time}. Features are extracted by putting each time series into different CNNs. After that, they concatenate those features together and put them into a new CNN framework. Large multivariate datasets are needed in order to train this deep architecture. While for our method, we focus on univariate time series and introduce two more branches that can extract multi-scale and multi-frequency information and further increase the prediction accuracy. \cite{dalto1deep} feeds CNN with variables post-processed using an input variable selection (IVS) algorithm. The key difference compared with \abbrev{} is that they aim at reducing the input size with different IVS algorithms. In contrast, we are exploring more raw information for CNN to discover.

In addition to classification, CNN is also used for time series metric learning. In~\cite{zheng2015convolutional}, Zheng~\emph{et al.} proposed a model called convolutional nonlinear neighbourhood components analysis that preforms CNN based metric learning and uses 1-NN as the classifier in the embedding space. 

Shapelets attract lots of attention because people can  detect shapes that are crucial to TSC, providing insights and interpretability. However, searching shapelets from all the time series segmentations is time consuming and some stoping methods are proposed to accelerate this procedure. In \cite{grabocka2014learning}, Grabocka~\emph{et al.} proposed a model that can learn global shapelets automatically instead of searching. As discussed in Section 3.2, \abbrev{} is general enough to be able to learn shapelets.  

CNN can achieve scale invariance to some extent by using the pooling operation. Thus, it is beneficial to introduce a multi-scale branch to extract short term as well as long term features. In image recognition,  CNNs keep feature maps in each stage and feed those feature maps altogether to the final fully connected layer~\cite{li2012multi} . By doing this, both short term and higher level features are preserved. For our model, we down sample the raw data into different time scales which provides low level features of different scales and higher level features at the same time.


\section{Experimental Results}
In this section, we conduct extensive experiments on various benchmark datasets to evaluate \abbrev{} and  compare it against many leading TSC methods. We have made an effort to include the most recent works.

\subsection{Experimental  setup}
We first describe the setup for our experiments.

\vspace*{0.1in}
\textbf{Baseline methods.}
For comprehensive evaluation, we evaluate two classical baseline methods: 1-NN with Euclidean distance (ED)~\cite{faloutsos1994fast} and 1-NN DTW~\cite{berndt1994using}. We also select 11 existing methods with state-of-the-art results published within the recent three years, including: DTW with a warping window constraint set through cross validation (DTW CV)~\cite{rakthanmanon2012searching}, Fast Shapelet (FS)~\cite{rakthanmanon2013fast} , SAX with vector space model (SV)~\cite{senin2013sax}, Bag-of-SFA-Symbols (BOSS)~\cite{schafer2015boss},  Shotgun Classifier (SC)~\cite{schafer2014towards}, time series based on a bag-of-features (TSBF)~\cite{baydogan2013bag},  Elastic Ensemble (PROP)~\cite{lines2015time},  1-NN Bag-Of-SFA-Symbols in Vector Space (BOSSVS)~\cite{schafer2015scalable}, Learn Shapelets Model(LTS)~\cite{grabocka2014learning}, and the Shapelet Ensemble (SE) model~\cite{bagnalltime}.

We also test standard convolutional neural network with the same number of parameters as in \abbrev{} to show the benefit of using the proposed multi-scale transformations and local convolution. For reference, we also list the results of \textbf{flat-COTE} (COTE), an ensemble model proposed by \emph{Bagnall et al.}~\cite{bagnalltime}, which uses the weighted votes over 35 different classifiers. \abbrev{} is orthogonal to flat-COTE and can be incorporated as a constituent classifier.

\vspace*{0.1in}
\textbf{Datasets.}
We evaluate all methods thoroughly on the UCR time series classification archive~\cite{UCRArchive}, which consists of 46 datasets selected from various real-world domains. We omit Car and Plane because a large portion of baseline methods do not provide related results. All the datasets in the archive are publicly available\footnote{\url{http://www.cs.ucr.edu/~eamonn/time_series_data/}}. Following the suggestions in \cite{paparrizos2015k}, we \emph{z-normalize} the following datasets during preprocessing: Beef, Coffee, Fish, OSULeaf and OliveOil.

All the experiments use the default training and testing set splits provided by UCR, and the results are rounded to three decimal places. For authoritative comparison, we adopt the experimental results collected by Bagnall~\emph{et al.}~\cite{bagnalltime} and Schafer~\cite{schafer2015scalable} for the baseline methods.

\begin{figure}[t]
    \centering
    \includegraphics[width=0.4\textwidth]{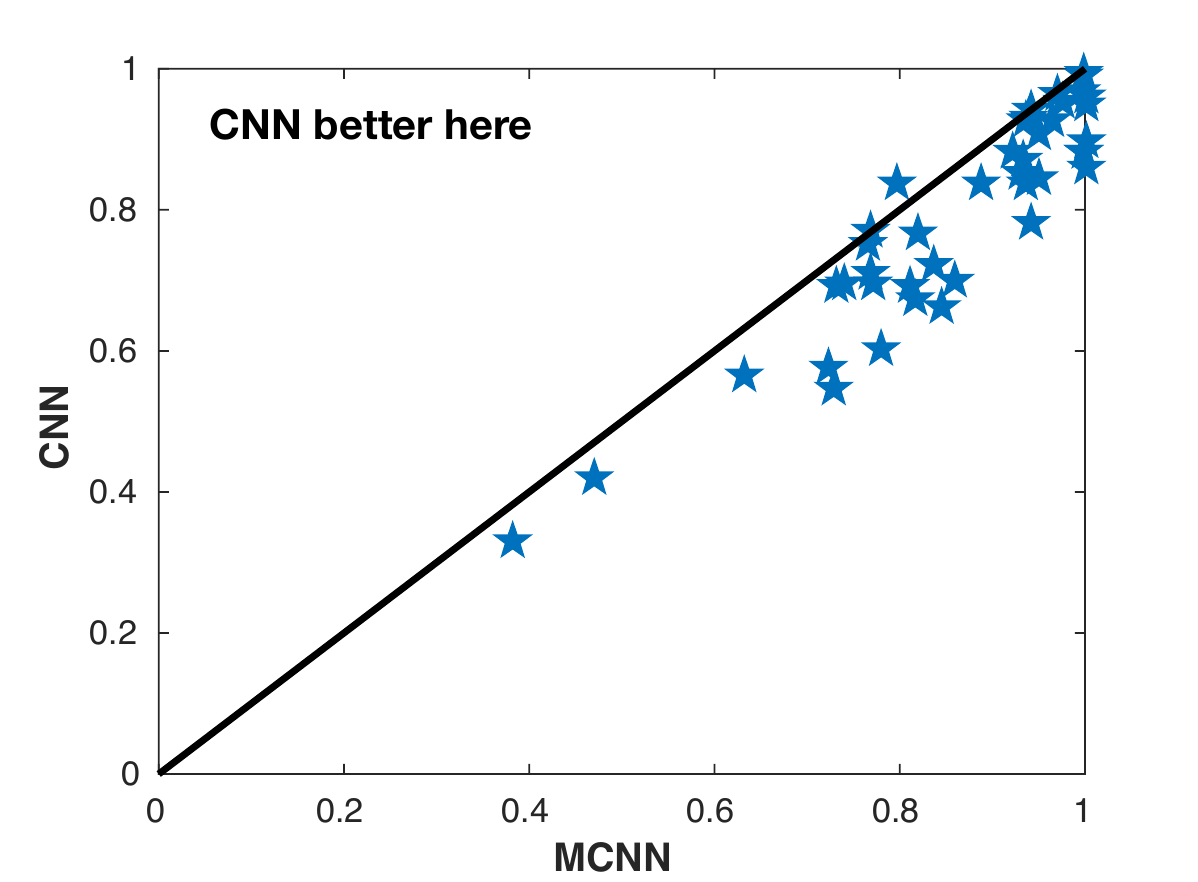}
    \caption{Scatter plot of test accuracies of standard CNN against \abbrev{} on all 44 UCR datasets. MCNN is better in all but 3 datasets.}
    \label{fig.cnn-mscnn}
\end{figure}

\vspace*{0.1in}
\textbf{Configuring MCNN.}
For \abbrev{}, we conduct the experiments on all the datasets with the same network architecture as in Figure 1. Since most of the datasets in the UCR archive are not large enough, we first use window slicing to increasing the size of the training size. For window slicing, we set the length of slices to be $0.9n$ where $n$ is the original length of the time series.  We set the number of filters to be 256 for the convolutional layers and include 256 neurons in the fully connected layer.

We use mini-batch stochastic gradient with momentum to update parameters in \abbrev{}. We adopt the \emph{grid search} for hyper-parameter tuning based on cross validation. The hyper parameters \abbrev{} include the filter size, pooling factor, and batch size.

In particular, the search space for the filter size is $\{0.05, 0.1, 0.2\}$,  which denotes the ratio of the filter length to the original time series length;  the search space for the pooling factor is $\{2,3,5\}$, which denotes the number of outputs of max-pooling. Early stopping is applied for preventing overfitting. Specifically, we use the error on the validation set to determine the best model. When the validation error does not get reduced for a number of epochs, the training terminates.

\abbrev{} is implemented based on theano~\cite{bergstra2010theano} and run on NVIDIA GTX TITAN graphics cards with 2688 cores and 6 GB global memory. For full replicability of the experiments, we will release our code and make it available in public\footnote{Source codes of the programs developed by our lab are published at \url{http://www.cse.wustl.edu/~ychen/psd.htm}.}.

\vspace*{0.1in}
\textbf{CNN vs. MCNN.}
Before comparing against other TSC classifiers, we first compare \abbrev{} with standard CNN. We test a CNN that has the same architecture and number of parameters as our \abbrev{} but does not have the multi-scale transformation and local convolutions.
Figure \ref{fig.cnn-mscnn} shows the scatter plot of the test accuracies of CNN and \abbrev{} on the 44 datasets.  We can see that \abbrev{} achieves better results on 41 out of 44 datasets. A binomial test confirms that \abbrev{} is significantly better than CNN at the 1\% level.

\begin{table*} [tp]
\caption{Testing error and rank for 44 ucr time series dataset.}
\label{tab.result}
\centering
\setlength\tabcolsep{1pt}
\begin{tabular}{|l|ccccccccccccccc|}
\hline
Dataset                 & DTW   & ED    & DTWCV & FS    & SV    & BOSS  & SE1   & TSBF  & TSF   & BOSSVS & PROP  & LS    & SE    & COTE  & MCNN  \\
\hline
Adiac            & 0.396    & 0.389   & 0.389       & 0.514          & 0.417   & \textbf{0.22}      & 0.373            & 0.245 & 0.261 & 0.302   & 0.353                   & 0.437                  & 0.435 & 0.233& 0.231 \\
Beef             & 0.367    & 0.467   & 0.333       & 0.447          & 0.467   & 0.2       & \textbf{0.133}            & 0.287 & 0.3   & 0.267   & 0.367                   & 0.24                   & 0.167 & \textbf{0.133}& 0.367 \\
CBF              & 0.003    & 0.148   & 0.006       & 0.053          & 0.007   & \textbf{0}         & 0.01             & 0.009 & 0.039 & 0.001   & 0.002                   & 0.006                  & 0.003 & 0.001& 0.002 \\
ChlorineCon      & 0.352    & 0.35    & 0.35        & 0.417          & 0.334   & 0.34      & 0.312            & 0.336 & 0.26  & 0.345   & 0.36                    & 0.349                  & 0.3   & 0.314& \textbf{0.203} \\
CinCECGTorso     & 0.349    & 0.103   & 0.07        & 0.174          & 0.344   & 0.125     & \textbf{0.021}            & 0.262 & 0.069 & 0.13    & 0.062                   & 0.167                  & 0.154 & 0.064& 0.058 \\
Coffee           & \textbf{0}        & \textbf{0}       & \textbf{0}           & 0.068          & \textbf{0}       & \textbf{0}         & \textbf{0  }              & 0.004 & 0.071 & 0.036   & \textbf{0 }                      & \textbf{0 }                     & \textbf{0} & \textbf{0}    & 0.036 \\
CricketX         & 0.246    & 0.423   & 0.228       & 0.527          & 0.308   & 0.259     & 0.297            & 0.278 & 0.287 & 0.346   & 0.203                   & 0.209                  & 0.218 & \textbf{0.154} &0.182 \\
CricketY         & 0.256    & 0.433   & 0.238       & 0.505          & 0.318   & 0.208     & 0.326            & 0.259 & 0.2   & 0.328   & 0.156                   & 0.249                  & 0.236 & 0.167 & \textbf{0.154} \\
CricketZ         & 0.246    & 0.413   & 0.254       & 0.547          & 0.297   & 0.246     & 0.277            & 0.263 & 0.239 & 0.313   & 0.156                   & 0.2                    & 0.228 & \textbf{0.128} & 0.142 \\
DiatomSizeR      & 0.033    & 0.065   & 0.065       & 0.117          & 0.121   & 0.046     & 0.069            & 0.126 & 0.101 & 0.036   & 0.059                   & 0.033                  & 0.124 & 0.082 & \textbf{0.023} \\
ECGFiveDays      & 0.232    & 0.203   & 0.203       & 0.004          & 0.003   & \textbf{0}         & 0.055            & 0.183 & 0.07  & \textbf{0 }      & 0.178                   & \textbf{0 }                     & 0.001 & \textbf{0} & \textbf{0 }    \\
FaceAll          & 0.192    & 0.286   & 0.192       & 0.411          & 0.244   & 0.21      & 0.247            & 0.234 & 0.231 & 0.241   & 0.152                  & 0.217                  & 0.263 & \textbf{0.105} & 0.235 \\
FaceFour         & 0.17     & 0.216   & 0.114       & 0.09           & 0.114   & \textbf{0}         & 0.034            & 0.051 & 0.034 & 0.034   & 0.091                   & 0.048                  & 0.057 & 0.091 & \textbf{0}     \\
FacesUCR         & 0.095    & 0.231   & 0.088       & 0.328          & 0.1     & \textbf{0.042}     & 0.079            & 0.09  & 0.109 & 0.103   & 0.063                   & 0.059                  & 0.087 & 0.057 & 0.063 \\
fiftywords       & 0.31     & 0.369   & 0.235       & 0.489          & 0.374   & 0.301     & 0.288            & 0.209 & 0.277 & 0.367   & \textbf{0.18}                    & 0.232                  & 0.281 & 0.191 & 0.19  \\
fish             & 0.177    & 0.217   & 0.154       & 0.197          & 0.017   & \textbf{0.011 }    & 0.057            & 0.08  & 0.154 & 0.017   & 0.034                   & 0.066                  & 0.023 & 0.029 & 0.051 \\
GunPoint         & 0.093    & 0.087   & 0.087       & 0.061          & 0.013   & \textbf{0}         & 0.06             & 0.011 & 0.047 & \textbf{0 }      & 0.007                   & \textbf{0 }                     & 0.02  & 0.007 & \textbf{0 }    \\
Haptics          & 0.623    & 0.63    & 0.588       & 0.616          & 0.575   & 0.536     & 0.607            & 0.488 & 0.565 & 0.584   & 0.584                   & 0.532                  & 0.523 & \textbf{0.488} & 0.53  \\
InlineSkate      & 0.616    & 0.658   & 0.613       & 0.734          & 0.593   & \textbf{0.511}     & 0.653            & 0.603 & 0.675 & 0.573   & 0.567                   & 0.573                  & 0.615 & 0.551 & 0.618 \\
ItalyPower       & 0.05     & 0.045   & 0.045       & 0.095          & 0.089   & 0.053     & 0.053            & 0.096 & 0.033 & 0.086   & 0.039                   & \textbf{0.03 }                  & 0.048 & 0.036 & \textbf{0.03}  \\
Lightning2       & 0.131    & 0.246   & 0.131       & 0.295          & 0.23    & 0.148     & \textbf{0.098 }           & 0.257 & 0.18  & 0.262   & 0.115                   & 0.177                  & 0.344 & 0.164 & 0.164 \\
Lightning7       & 0.274    & 0.425   & 0.288       & 0.403          & 0.342   & 0.342     & 0.274            & 0.262 & 0.263 & 0.288   & 0.233                   & \textbf{0.197 }                 & 0.26  & 0.247 & 0.219 \\
MALLAT           & 0.066    & 0.086   & 0.086       & \textbf{0.033}          & 0.199   & 0.058     & 0.092            & 0.037 & 0.072 & 0.064   & 0.05                    & 0.046                  & 0.06  & 0.036 & 0.057 \\
MedicalImages    & 0.263    & 0.316   & 0.253       & 0.433          & 0.516   & 0.288     & 0.305            & 0.269 & \textbf{0.232} & 0.474   & 0.245                   & 0.27                   & 0.396 & 0.258& 0.26  \\
MoteStrain       & 0.165    & 0.121   & 0.134       & 0.217          & 0.117   & \textbf{0.073 }    & 0.113            & 0.135 & 0.118 & 0.115   & 0.114                   & 0.087                  & 0.109 & 0.085 & 0.079 \\
NonInvThorax1    & 0.21     & 0.171   & 0.189       & 0.171          &         & 0.161     & 0.174            & 0.138 & 0.103 & 0.169   & 0.178                   & 0.131                  & 0.1   & 0.093 & \textbf{0.064 }\\
NonInvThorax2    & 0.135    & 0.12    & 0.12        & 0.12           &         & 0.101     & 0.118            & 0.13  & 0.094 & 0.118   & 0.112                   & 0.089                  & 0.097 & 0.073 & \textbf{0.06 } \\
OliveOil         & 0.167    & 0.133   & 0.133       & 0.213          & 0.133   & 0.1       & 0.133            & \textbf{0.09}  & 0.1   & 0.133   & 0.133                   & 0.56                   & 0.1   & 0.1 & 0.133 \\
OSULeaf          & 0.409    & 0.483   & 0.388       & 0.359          & 0.153   & \textbf{0.012}     & 0.273            & 0.329 & 0.426 & 0.074   & 0.194                   & 0.182                  & 0.285 & 0.145 & 0.271 \\
SonyAIBORobot    & 0.275    & 0.305   & 0.304       & 0.314          & 0.306   & 0.321     & 0.238            & 0.175 & 0.235 & 0.265   & 0.293                   & 0.103                  & \textbf{0.067} & 0.146 & 0.23  \\
SonyAIBORobotII  & 0.169    & 0.141   & 0.141       & 0.215          & 0.126   & 0.098     & \textbf{0.066 }           & 0.196 & 0.177 & 0.188   & 0.124                   & 0.082                  & 0.115 & 0.076 & 0.07  \\
StarLightCurves  & 0.093    & 0.151   & 0.095       & 0.06           & 0.108   & \textbf{0.021}     & 0.093            & 0.022 & 0.036 & 0.096   & 0.079                   & 0.033                  & 0.024 & 0.031 & 0.023 \\
SwedishLeaf      & 0.208    & 0.213   & 0.154       & 0.269          & 0.275   & 0.072     & 0.12             & 0.075 & 0.109 & 0.141   & 0.085                   & 0.087                  & 0.093 & \textbf{0.046} & 0.066\\
Symbols          & 0.05     & 0.1     & 0.062       & 0.068          & 0.089   & 0.032     & 0.083            & 0.034 & 0.121 & \textbf{0.029}   & 0.049                   & 0.036                  & 0.114 & 0.046 & 0.049 \\
SyntheticControl & 0.007    & 0.12    & 0.017       & 0.081          & 0.013   & 0.03      & 0.033            & 0.008 & 0.023 & 0.04    & 0.01                    & 0.007                  & 0.017 & \textbf{0} & 0.003 \\
Trace            & \textbf{0 }       & 0.24    & 0.01        & 0.002          & \textbf{0}       & \textbf{0 }        & 0.05             & 0.02  & \textbf{0}     & \textbf{0}       & 0.01                    & \textbf{0  }                    & 0.02  & 0.01 & \textbf{0}     \\
TwoLeadECG       & \textbf{0}        & 0.09    & 0.002       & 0.113          & 0.004   & 0.004     & 0.029            & 0.001 & 0.112 & 0.015   & \textbf{0 }                      & 0.003                  & 0.004 & 0.015 & 0.001 \\
TwoPatterns      & 0.096    & 0.253   & 0.132       & 0.09           & 0.011   & 0.016     & 0.048            & 0.046 & 0.053 & 0.001   & 0.067                   & 0.003                  & 0.059 & \textbf{0} & 0.002 \\
UWaveX           & 0.272    & 0.261   & 0.227       & 0.293          & 0.324   & 0.241     & 0.248            & \textbf{0.164} & 0.213 & 0.27    & 0.199                   & 0.2                    & 0.216 & 0.196 & 0.18  \\
UWaveY           & 0.366    & 0.338   & 0.301       & 0.392          & 0.364   & 0.313     & 0.322            & \textbf{0.249 }& 0.288 & 0.364   & 0.283                   & 0.287                  & 0.303 & 0.267 & 0.268 \\
UWaveZ           & 0.342    & 0.35    & 0.322       & 0.364          & 0.357   & 0.312     & 0.346            & \textbf{0.217} & 0.267 & 0.336   & 0.29                    & 0.268                  & 0.273 & 0.265 & 0.232 \\
wafer            & 0.02     & 0.005   & 0.005       & 0.004          & 0.002   & \textbf{0.001}     & 0.002            & 0.004 & 0.047 & \textbf{0.001}   & 0.003                   & 0.004                  & 0.002 & \textbf{0.001} & 0.002 \\
WordSynonyms     & 0.351    & 0.382   & 0.252       & 0.563          & 0.436   & 0.345     & 0.357            & 0.302 & 0.381 & 0.439   & \textbf{0.226}                   & 0.34                   & 0.403 & 0.266 & 0.276 \\
youga            & 0.164    & 0.17    & 0.156       & 0.249          & 0.151   & \textbf{0.081 }    & 0.159            & 0.149 & 0.157 & 0.169   & 0.121                   & 0.15                   & 0.195 & 0.113 & 0.112
\\
\hline
\#best & 3& 1& 1& 1& 2& 15 & 5& 4& 2& 5& 4& 6& 2& 11& 10\\
rank mean &10.05 & 12.32 & 9.01 & 12.88 & 10.25 & 5.40 & 8.76 & 7.43 & 8.27 & 8.61 & 6.00 & 5.63 & 7.61 & 3.61 & 3.95\\
\hline
\end{tabular}
\end{table*}

\subsection{Comprehensive evaluation}

Table~\ref{tab.result} shows a comprehensive evaluation of all methods on the UCR datasets. For each dataset, we rank all the 15 classifiers. The last row of  Table~\ref{tab.result}  shows the mean rank for each solver (lower is better). We see that \abbrev{} is very competitive,  achieving the highest accuracy on 10 datasets. \abbrev{} has a mean rank of 3.95, lower than all the state-of-the-art methods except for COTE, which is an ensemble of 35 classifiers.

To further analyze the performance, we make pairwise comparison for each algorithm against \abbrev{}.  Binomial test (BT) and the Wilcoxon signed rank test (WSR) are used to measure the significance of difference. Corresponding $p$-values are listed in table~\ref{tab.pairwise-comp}, indicating that \abbrev{} is significantly better than all the other methods except for BOSS and COTE at the 1\% level ($p < 0.01$). Moreover, it shows that the differences between COTE, BOSS, and \abbrev{} are not significant.

\begin{table} [t]
\caption{Pairwise comparison with \abbrev{}. $p$(BT) and $p$(WSR) are the $p$-values of binomial test and Wilcoxon signed rank test, respectively.}
\label{tab.pairwise-comp}
\centering
\setlength\tabcolsep{2pt}
\begin{tabular}{|l|c|c|c|c|c|}
\hline
Model & \#better & \#tie & \#worse & $p$(BT) & $p$(WSR) \\
\hline
DTW & 5 & 2 & 37 & $9.43 \times 10^{-7}$ &$4.24 \times 10^{-7}$\\
\hline
ED & 1 & 1 & 42 & $2.15 \times 10^{-10}$&$1.54 \times 10^{-8}$\\
\hline
DTWCV & 7 & 1 & 36 &$8.96 \times 10^{-9}$&$3.24 \times 10^{-6}$\\
\hline
FS & 1 & 0 & 43& $5.12 \times 10^{-12}$&$1.11 \times 10^{-8}$\\
\hline
SV &4 & 3 & 35& $3.35 \times 10^{-7}$&$2.48	 \times 10^{-6}$\\
\hline

SC & 5 & 2 & 37 &$4.43 \times 10^{-7}$&$1.60 \times 10^{-5}$\\
\hline
TSBF & 13 & 1 & 30 & $1.73 \times 10^{-2}$&$1.60 \times 10^{-3}$\\
\hline
TSF & 4 & 1 & 39 & $3.10 \times 10^{-8}$ & $5.08 \times 10^{-7}$\\
\hline
BOSSVS & 8 & 5 & 31 & $2.94 \times 10^{-4}$&$6.77 \times 10^{-5}$\\
\hline
SE & 7 & 1 & 36 & $8.96 \times 10^{-6}$&$8.90 \times 10^{-5}$\\
\hline
PROP & 11 & 5 & 28 & $9.50 \times 10^{-3}$&$1.20 \times 10^{-2}$\\
\hline
LTS & 10 & 4 & 30 & $2.20 \times 10^{-3}$&$1.60 \times 10^{-2}$\\
\hline
BOSS & 16 & 4 & 24 & $2.68 \times 10^{-1}$&$7.38 \times 10^{-2}$\\
\hline
COTE & 22 & 2 & 20 & $8.78 \times 10^{-1}$ & $5.95 \times 10^{-1}$\\
\hline

\end{tabular}
\end{table}

\begin{figure*}[t]
    \centering
    \includegraphics[width=1\textwidth]{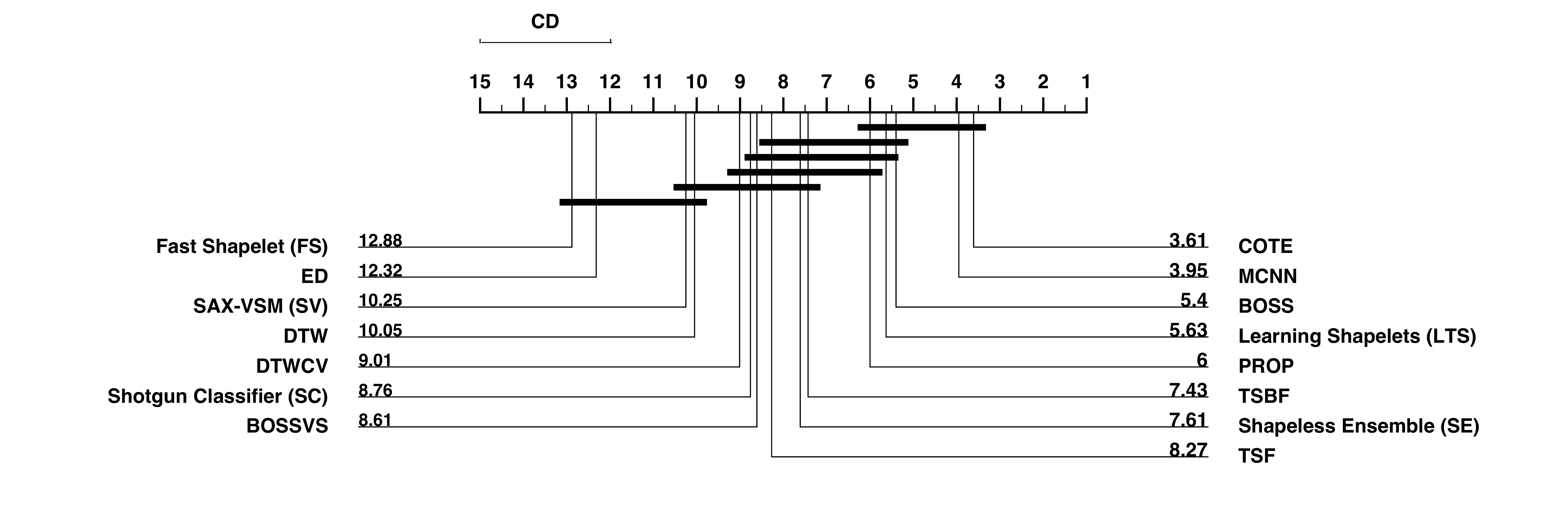}
    \caption{Critical Difference Diagram~\cite{demvsar2006statistical} over the mean ranks of \abbrev{}, 13 baseline methods and the COTE ensemble. The critical difference is 3.01. COTE is the best classifier which ensembles 35 classifiers. \abbrev{} performs equally well compared with COTE.}
    \label{fig.cd}
\end{figure*}

Figure \ref{fig.cd} shows the critical difference diagram, as proposed in \cite{demvsar2006statistical}. The values shown on the figure are the average rank of each classifier. Bold lines indicate groups of classifiers which are not significantly different. The critical difference (CD) length is shown on the graph. Figure \ref{fig.cd} is evaluated on \abbrev{}, all baseline methods and COTE. \abbrev{} is among the most accurate classifiers and its performance is very close to COTE. It is quite remarkable that \abbrev{}, a single algorithm, obtains the same state-of-the-art performance as an ensemble model consisting of 35 different classifiers. Note that \abbrev{} is orthogonal to flat-COTE as \abbrev{} can also be included as a predictor in flat-COTE to further improve the performance. There are obvious margins between \abbrev{} and other baseline classifiers.

\begin{figure*}[t]
    \centering
    \includegraphics[width=1\textwidth]{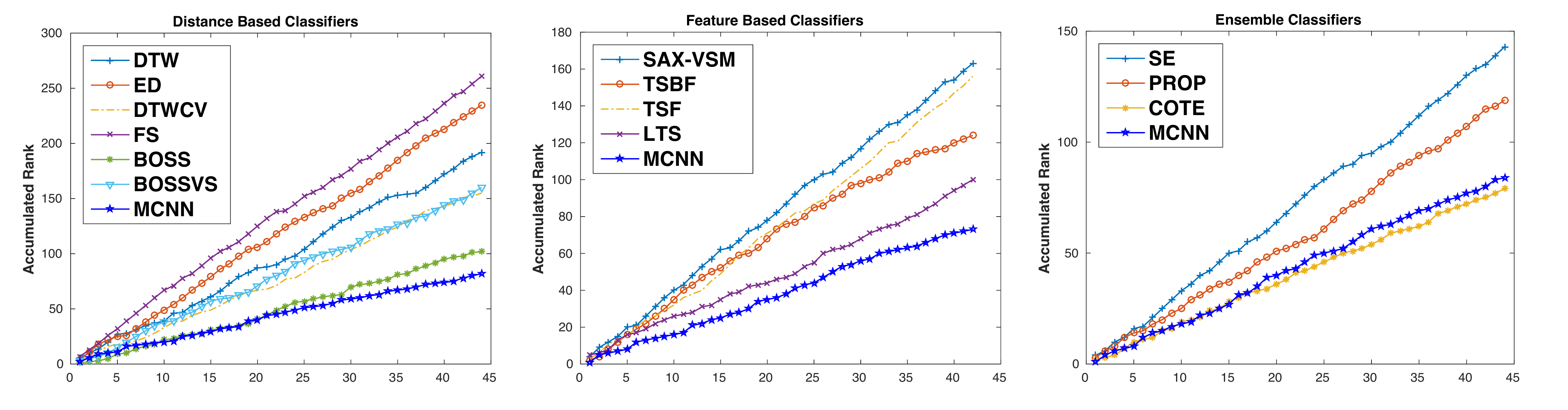}
    \caption{Comparison of \abbrev{} against three groups of classifiers in terms of accumulated ranks.}
    \label{fig.acc-rank}
\end{figure*}

We now group these classifiers into three categories and provide mode detailed analysis.

\vspace*{0.1in}
\textbf{Distance based classifiers.}
These classifiers use nearest neighbor algorithms based on distance measures between time series. The simplest distance measure is the Euclidean distance (ED). Dynamic time warping (DTW) is proposed to extract the global similarity while addressing the phase shift problem. DTW with 1-NN classifier has been hard to beat for a long time and now become a benchmark method. DTW with warping set through cross-validation (DTWCV) is also a traditional bench mark. k-NN classifiers also uses transformed features. Fast shapelet (FS) search shapelet on a lower transformed space. Bag-of-SFA-Symbols (BOSS)~\cite{schafer2015boss} proposes a distance based on histograms of symbolic Fourier approximation words. The BOSSVS model combines the BOSS model with the vector space model to reduce the time complexity. They are all combine with 1-NN for final prediction.

By grouping distance based classifiers together, we can  compare their average performance with \abbrev{}. In order to illustrate the overall performance  of different algorithms, we plot accumulated rank on all the tested datasets in Figure \ref{fig.acc-rank}. We order all the datasets alphabetically by  their name and show the accumulated rank. For example, if a method is always ranked \#1, its accumulated rank is $N$ for the $N^{th}$ dataset. From Figure \ref{fig.acc-rank}, we see that \abbrev{} has the lowest accumulated rank, outperforming all the distance based classifiers.

\vspace*{0.1in}
\textbf{Feature based classifiers.}
For feature based classifiers, we selected SAX-VSM, TSF, TSBF, LTS. Symbolic aggregate approximation (SAX) has become a classical method to discretize time series based on piecewise mean value., SAX-VSM achieved state-of-the-art classification accuracy on UCR dataset by combining vector space Model (VSM) with SAX.
Time series forest (TSF) divide time series into different intervals and calculate the mean, standard deviation and slop as interval features. Instead of using traditional entropy gain, TSF proposed a new split criteria by adding an addition term measuring the nearest distance between interval features and split threashold to the entropy and achieved better results than traditional random forests.
The bag-of-features framework (TSBF) also extracts interval features with different scales. The features from each interval form an instance, and each time series forms a bag. Random forest is used to build a supervised codebook and classify time series. Finally, learning time series shapelets (LTS) provides not only competitive results, but also the ability to learn shapelets directly. Classification is made based on logistic regression.

The middle plot of figure \ref{fig.acc-rank} compares the performance of \abbrev{} against some on feature based classifiers, including SV, TSBF, TSF, and LTS. It is clearly that \abbrev{} is substantially better than these feature based classifiers, as its accumulated rank is consistently the lowest by a large margin.

\vspace*{0.1in}
\textbf{Ensemble based classifiers}
There is a growing trend in ensembling different classifiers together to achieve higher accuracy. The Elastic Ensemble (PROP) combined 11 distinct classifiers based on elastic distance measures  through a weighted ensemble scheme. This was the first classifier that significantly outperformed DTW at that time. Shapelet ensemble (SE) combines shapelet transformations with a heterogeneous ensemble method. The weight of each classifier is assigned based on the cross validation accuracy. The flat collective of transform-based ensembles (flat-COTE) is an ensemble of 35 different classifiers based on features from time and frequency domains and has achieved state-of-the-art accuracy performance. Despite its high testing accuracy, ensemble methods suffer high complexity during the training process as well as testing. From the third plot in figure \ref{fig.acc-rank}, we can observe that \abbrev{} is very close to COTE and much better than SE and PROP. Critical difference analysis in Figure~\ref{fig.cd} also confirms that there is no significant difference between COTE and \abbrev{}.
It is in fact quite remarkable that a single algorithm in \abbrev{} can match the performance of the COTE ensemble. The performance of \abbrev{} is likely to improve further if it is trained with larger datasets, since convolutional neural networks are known to be able to absorb huge training data and make improvements.


\section{Conclusions}
We have presented \name (\abbrev{}), a convolutional neural network tailored for time series classification.
\abbrev{} unifies feature extraction and classification, and jointly learns the parameters through back propagation. It leverages the strength of CNN to automatically learn good feature representations in both time and frequency domains.
In particular, \abbrev{} contains multiple branches that perform various transformations of the time series, which extract features of different frequency and time scales, addressing the limitation of many previous works that they only extract features at a single time scale.
We have also discussed the insights that learning convolution filters in \abbrev{}  generalizes shapelet learning, which in part explains the excellent performance of \abbrev{}.

We have conducted comprehensive experiments and compared with leading time series classification models. We have demonstrated that \abbrev{} achieves state-of-the-art performance and outperforms many existing models by a large margin, especially when enough training data is present. 

More importantly, an advantage of CNNs is that they can absorb massive amount of data to learn good feature representations. Currently, all the TSC datasets we have access to are not very large, ranging from a training size of around 50 to a few thousands. We envision that \abbrev{} will show even greater advantages in the future when trained with much larger datasets.
We hope \abbrev{} will inspire more research on integrating deep learning  with time series data analysis. For  future work, we will investigate how to augment \abbrev{} for time series classification by incorporating other side information from multiple sources, such as text, image and speech.

\section{Acknowledgments}
The authors are supported in part by the IIS-1343896, DBI-1356669, and III-1526012 grants from the National Science Foundation of the United States, a Microsoft Research New Faculty Fellowship, and a Barnes-Jewish Hospital Foundation grant.

%
\bibliographystyle{abbrv}

\bibliography{Chen,timeseries}  
%
%

\end{document}